\documentclass{article}

\PassOptionsToPackage{numbers, compress}{natbib}

\usepackage[preprint]{neurips_2022}

\usepackage[utf8]{inputenc} 
\usepackage[T1]{fontenc}    
\usepackage{hyperref}       
\hypersetup{
    colorlinks,
    linkcolor={red!50!black},
    citecolor={blue!50!black},
    urlcolor={blue!80!black}
}
\usepackage{url}            
\usepackage{booktabs}       
\usepackage{amsfonts}       
\usepackage{nicefrac}       
\usepackage{microtype}      
\usepackage{xcolor}         
\usepackage{graphicx} 
\usepackage{multirow}
\usepackage{enumitem}
\usepackage{flafter}
\usepackage{listings}
\usepackage{caption}
\usepackage{subcaption}

\title{OpenAL: Evaluation and Interpretation of Active Learning Strategies}

\author{%
  William Jonas\\
  Dataiku\\
  \texttt{williamjonas@hotmail.fr} \\
  \And
  Alexandre Abraham\\
  Dataiku\\
  \texttt{abraham.alexandre@gmail.com} \\
  \And
  Léo Dreyfus-Schmidt\\
  Dataiku\\
  \texttt{leo.dreyfus-schmidt@dataiku.com} \\
}

\usepackage{todonotes}
\usepackage{soul}

\makeatletter
 \if@todonotes@disabled
 
 \else
 
 \fi
 \makeatother

\usepackage{xspace}

\makeatletter
\DeclareRobustCommand\onedot{\futurelet\@let@token\@onedot}
\def\@onedot{\ifx\@let@token.\else.\null\fi\xspace}

\def\ie{\emph{i.e}\onedot}

\makeatother

\colorlet{hardcolor}{red!30}
\colorlet{mediumcolor}{orange!30}
\colorlet{easycolor}{green!30}

\usepackage{ulem}
\makeatletter
\newcommand\reformulate[1][red]{\bgroup \markoverwith{\lower3.5\p@\hbox{\sixly \textcolor{#1}{\char58}}}\ULon}
\font\sixly=lasy6 
\makeatother
\begin{document}





\maketitle

\begin{abstract}

Despite the vast body of literature on Active Learning (AL), there is no comprehensive and open benchmark allowing for efficient and simple comparison of proposed samplers. Additionally, the variability in experimental settings across the literature makes it difficult to choose a sampling strategy, which is critical due to the one-off nature of AL experiments. To address those limitations, we introduce OpenAL, a flexible and open-source framework to easily run and compare sampling AL strategies on a collection of realistic tasks. The proposed benchmark is augmented with interpretability metrics and statistical analysis methods to understand when and why some samplers outperform others. Last but not least, practitioners can easily extend the benchmark by submitting their own AL samplers.
\end{abstract}

\section{Introduction}



Active Learning (AL) has proved its worth in practice to optimize labeling tasks~\cite{ren2021survey}. However, it remains challenging to apply in practice
as its benefit can vary significantly depending on the task~\cite{lowell2018practical}. The optimal sampler may depend on several experimental hyperparameters, such as the initial labeled set size, the batch size, the ML model used, or the number of iterations, among others. Those hyperparameters values vary substantially between studies, even for similar tasks, as shown in Table~\ref{other-papers-al-settings-table}.

\begin{table}
  \caption{AL experiment parameters}
  \label{other-papers-al-settings-table}
  \centering
  \begin{tabular}{lllll}
    \toprule
    Paper   &   Dataset &   Init size   &   Batch size  &   nb iterations    \\
    \midrule
    \multirow{3}{5.5cm}{Active Learning for convolutional neural networks: a core set approach~\cite{sener2017active}}  &   CIFAR 10    &  10\% &   10\% &   3      \\  
        &   CIFAR 100   &   10\% &   10\% &   3 \\
        &   SVHN    &   1\%  &   8\% then 43\% &   3    \\
    \midrule 
    \multirow{3}{5.5cm}{Deep batch active learning by diverse, uncertain gradient lower bounds~\cite{ash2019deep}}   & SVHN    &   100 &   100 &   350  \\
        &   OpenML \#156 &   100 &   1000    &   4    \\
        &   CIFAR 10    &   100 &   10000   &   4    \\
    \midrule 
    \multirow{4}{5.5cm}{Variational Adversarial Active Learning \cite{sinha2019variational}}  &   CIFAR 10    &   10\% &   5\%  &   6   \\
        &   CIFAR 100    &   10\% &   5\%  &   6   \\
        &   Caltech-256    &   10\% &   5\%  &   6   \\
        &   ImageNet    &   10\% &   5\%  &   6   \\
    \midrule 
    \multirow{3}{5.5cm}{BatchBALD: Efficient and Diverse Batch Acquisition for Deep Bayesian Active Learning~\cite{kirsch2019batchbald}}    &   MNIST   &   10  &   10  &   25  \\
        &   EMNIST   &   10  &   10  &   25  \\
        &   CINIC-10    &   200 &   10  &   120 \\
    \bottomrule
  \end{tabular}
\end{table}




 This diversity in experimental settings impairs reproducibility and makes methods comparisons arduous. Existing AL benchmarks have tackled this variability by fixing some parameters arbitrarily or targeting specific AL problems, such as using only Logistic Regression as a base learner~\cite{yang2018benchmark}, outlier detection~\cite{trittenbach2021overview}, or structural reliability~\cite{moustapha2022active}. But comparing sampling strategies reliably requires to repeat the experiments several times~\cite{kottke2017challenges} using various tasks and models~\cite{munjal2020towards}. OpenAL follows those best practices and encompasses various realistic tasks, models, and use cases. We designed them as close as possible to real tasks. We address the following caveats:

\begin{itemize}[left=4pt,label={}]

\item{\bf Initialization induced variability.} It has been proven that the variance in performance induced by the initial set of selected samples can be greater than the difference between sampling strategies~\cite{kottke2017challenges}. We propose to use a 10-fold stratified shuffle split to get enough significance when comparing methods~\cite{dietterich1998approximate}.

\item{\bf Plausibility of the experimental setting.} Research task settings must be well representative of real-life ones to be helpful. Experiments on CIFAR-10 in the literature often vary from 6 batches of 5\% of the whole dataset to 3 batches of 10\%~\cite{sener2017active}. According to earlier work on realistic applications of AL~\cite{wiens2010active}, it is usually used to reduce data labeling between 1\% and 10\%. 
OpenAL's default is to label 1\% of the data in 10 iterations on tabular and image classification tasks. We kickstarted the image classification models using transfer learning or self-supervision, following the industry best practices.

\item{\bf Reproducibility.} Our framework is open source, and all experiments results are made available and can be easily run again. We provide the accuracies and other AL metrics for the most common AL samplers, along with all train, test, and initial batch indices used for those experiments. 

\item{\bf Online evaluation of sampling strategies.} Research works rely on the area under the accuracy curve of a left-out test set to evaluate the performance of AL strategies. This testing set is not available in real experiments making it hard to trust their behavior online~\cite{kottke2019limitations}. OpenAL logs unsupervised metrics to improve the offline strategies' interpretability and be able to interpret their behavior online~\cite{abraham2020rebuilding}.


\end{itemize}






We first start by describing the setup of our tasks, the model selection methodology, and the evaluation criteria for sampling strategies. Then we present the results of our experiments per strategy across all tasks. We finish by focusing on the metrics observed and how they explain the performance of some strategies. Finally we open new perspectives on AL experiments and how this benchmark could be useful and extended in the future.

\section{Evaluation framework}



OpenAL features eleven classification tasks on tabular datasets and four on image datasets. 
Tabular datasets come from OpenML~\cite{vanschoren2014openml, feurer2021openml} and must be plausible enough \ie having at least 10000 samples to justify the cost of setting up an AL pipeline and being non-trivial, or not solvable easily with 1\% randomly selected samples. We were left with 11 tasks, which we deemed sufficient to obtain reliable results to compare AL strategies.




\textbf{Cross-validation.} Each task is repeated ten times with different test sets and batch initialization. We use a stratified shuffle split with 20\% of the data dedicated to the test. This amount of repetition is said to provide enough significance for method comparison~\cite{kottke2017challenges}. Our accuracy plots display confidence intervals of $10^{th}$ to $90^{th}$ quantiles over the ten folds.



\textbf{Active learning experimental setting.} We chose experimental parameters to be as close as possible to industrial use cases. Each experiment starts with 0.1\% randomly selected labeled data with at least one sample from each class. Nine iterations follow it with batch size 0.1\% to end up with a total of 1\% of the data labeled. We do not use a specific stopping criterion and stop the experiment when this labeling budget is exhausted. In most experiments, this budget allows the best AL method to reach a performance plateau, as shown in experiments in Section~\ref{section:experiments}. OpenAL includes seminal uncertainty-based strategies~\cite{settles2009active} (Margin, Confidence, and Entropy), weighted KMeans (WKMeans)~\cite{zhdanov2019diverse}, incremental weighted KMeans (IWKMeans)~\cite{abraham2021sample}, and k-center greedy (Kcenter)~\cite{sener2017active}. Note that what most literature works call core-sets use k-center greedy because of the latter's high computational cost. We call it by its original name to avoid any confusion. Since KCenter relies on the weights of the penultimate layer of a neural network for its computation, we used the embedding method proposed in scikit-learn for embedding tree models. It vectorizes the data using a PCA computed on the activation of the tree leaves.



\textbf{Selection of the best model.} Models are usually selected using cross-validation, which is tricky to perform in Active Learning where labeled data is scarce~\cite{limberg2020beyond}. We expect the practitioners to have prior knowledge of which models could perform well for the task at hand. For tabular datasets and MNIST, we simulate this prior knowledge by doing model selection over the whole dataset using a 5-fold cross-validation. We consider a multi-layer perceptron and two tree-based models, Random Forest and Gradient Boosting Tree, as they are known to excel on tabular data.
For CIFAR-10 and CIFAR-100, we use embeddings precomputed on ImageNet and finetune the last layer. For CIFAR-10 only, we also consider embeddings precomputed on unlabeled data using contrastive learning~\cite{chen2020simple}.
 

\textbf{Experiment caching for easy comparison.} All the benchmark elements are seeded, which guarantees reproducibility at the machine level. Because seeded number generation may change from one machine to another, we also provide the indices of all train and test indices used in our benchmark. Once a strategy has run, all its corresponding metrics results are cached and can be used for plotting or method comparison. Running a new strategy is as simple as taking the dedicated notebook, wrapping the strategy in our sampler formalism, and running it. Submitting the results can then be done through a GitHub pull request.


\section{Software}

OpenAL is coded in Python and available through the GitHub platform\footnote{\url{https://github.com/dataiku-research/OpenAL}}. We also provide documentation explaining how to install, use, and publish results using our framework\footnote{\url{https://dataiku-research.github.io/OpenAL/}}. The repository contains all results of previous experiences. Running the benchmark on all reference samplers or on a new one is as simple as 3 lines of code that are contained in the \texttt{main\_run.py} file:

\begin{lstlisting}[language=Python]
initial_conditions = load_initial_conditions(dataset_id)
experimental_parameters = load_experiment(dataset_id, initial_conditions)
run(experimental_parameters, methods)

\end{lstlisting}

All experiments are modular and split in blocks for easy running and customization.
\emph{Initial conditions} contain the samples initially labeled and the number of folds. \emph{Experimental parameters} include the batch size and the number of iterations. The \emph{run} function runs the experiment and generates accuracy and metrics results in a dedicated folder that the user can submit through a pull request for validation. After replicating the results on our side, we will integrate this new sampler into OpenAL and share the results with the community.

\section{Experiments and results}
\label{section:experiments}

The tasks included in OpenAL are listed in Table~\ref{dataset-table}.
We report accuracy and the following set of metrics measured during our experiments:
\begin{itemize}[left=4pt,label={}]
    \item{\bf Agreement.} Agreement ratio between the inductive model and a 1-nearest-neighbor (1-NN) classifier trained on labeled data. We expect a high agreement to be correlated with good exploration.
    \item{\bf Contradictions.} Ratio of test samples where the models at the previous iteration and current iteration disagree. It is an upper bound on accuracy change from one iteration to the other.
    \item{\bf Hard exploration.} Ratio of test samples where the 1-NN of the previous iteration and current iteration disagree.
    \item{\bf Top exploration.} Mean difference of the distance between test samples and their nearest neighbour in the labeled pool from one iteration to the next.
    \item{\bf Violations.} This unsupervised metric measures how many data compliance rules computed on the test set are violated in the labeled dataset~\cite{fariha2021conformance}. Conformance rules are computed by extracting eigenvectors on the reference dataset and setting conformance boundaries based on the standard deviation of the projected reference data. Sample conformance is given by the number of times its projections fall outside the conformance boundaries. Overall, the highest the violation, the more the labeled samples deviate from the test set.
    
    
\end{itemize}

\begin{table}
  \centering
  \begin{tabular}{lllll}
    \toprule
    Name     & \#samples     & \#classes  & \#features &   class balance \\
    \midrule
    \#1461 Bank-marketing    &   45221   &   2   &   7/9  &   0.88 / 0.12  \\
    \#1471 Eeg-eye-state     &   14980   &   2   &   14/0  &   0.55 / 0.45  \\
    \#1502 Skin-segmentation &   245057  &   2   &   3/0  &   0.21 / 0.79  \\
    \#1590 Adult             &   48842   &   2   &   6/8  &   0.76 / 0.24  \\
    \#40922 Run or walk information  &   88588   &   2   &   6/0  &   0.5 / 0.5   \\
    \#41138 APSFailure       &   76000   &   2   &   170/0  &   0.98 / 0.02   \\
    \#41162 Kick             &   72983   &   2   &   14/18  &   0.88 / 0.12   \\
    \#42395 Santander Customer Satisfaction  &   200000  &   2   &   200/0  &   0.9 / 0.1   \\
    \#42803 Road-safety      &   363243  &   3   &   61/5  &   0.66 / 0.29 / 0.05  \\
    \#43439 Medical-Appointment-No-Shows &   110527  &   2   &   8/4  &   0.8 / 0.2   \\
    \#43551 Employee-Turnover-at-TECHCO  &   34452   &   2   &   9/1  &   0.02 / 0.98  \\
    MNIST                   &   70000   &   10  &   28x28   &   0.1 each   \\
    CIFAR10                 &   60000   &   10  &   32x32x3   &   0.1 each   \\
    CIFAR100                &   60000   &   100 &   32x32x3   &   {\bf 0.01 each}   \\
    \bottomrule
  \end{tabular}
  \caption{OpenAL datasets' characteristics. For tabular data, features correspond ton continuous/categorical features. For images, the shape of one image is given.}
  \label{dataset-table}
\end{table}


\begin{table}
  \centering
  \begin{tabular}{rccccccc}
    \toprule
Dataset   &     Random        &     KMeans        &     Confidence    &     Margin        &     KCenter       &     WKmeans      \\
\midrule
1471      &     68.6 $\pm0.8$ &     68.7 $\pm1.2$ &     69.7 $\pm1.0$ &     69.7 $\pm1.0$ &     67.4 $\pm0.6$ & \bf 71.2 $\pm1.1$\\
41138     &     98.4 $\pm0.1$ &     98.4 $\pm0.1$ &     98.9 $\pm0.1$ &     98.9 $\pm0.1$ &     98.7 $\pm0.1$ & \bf 99.0 $\pm0.1$\\
1502      &     98.7 $\pm0.4$ &     99.3 $\pm0.0$ &     99.2 $\pm0.2$ &     99.2 $\pm0.2$ & \bf 99.5 $\pm0.1$ & \bf 99.5 $\pm0.1$\\
1590      &     81.8 $\pm1.0$ &     80.7 $\pm0.8$ &     81.5 $\pm1.0$ &     81.5 $\pm1.0$ & \bf 82.0 $\pm0.6$ & \bf 82.9 $\pm0.6$\\
41162     &     85.0 $\pm0.7$ &     84.3 $\pm0.9$ &     85.7 $\pm0.9$ &     85.7 $\pm0.9$ & \bf 86.7 $\pm0.5$ & \bf 86.3 $\pm0.8$\\
43439     &     76.6 $\pm0.3$ &     76.1 $\pm0.6$ &     76.5 $\pm0.4$ &     76.5 $\pm0.4$ & \bf 77.3 $\pm0.9$ & \bf 77.0 $\pm0.5$\\
40922     &     96.6 $\pm0.4$ &     96.0 $\pm0.6$ & \bf 97.7 $\pm0.5$ & \bf 97.7 $\pm0.5$ & \bf 97.3 $\pm0.1$ & \bf 97.7 $\pm0.5$\\
42395     &     89.7 $\pm0.1$ &     89.6 $\pm0.1$ & \bf 89.8 $\pm0.1$ & \bf 89.8 $\pm0.1$ & \bf 89.8 $\pm0.1$ & \bf 89.8 $\pm0.1$\\
43551     &     97.2 $\pm0.5$ &     95.2 $\pm1.8$ & \bf 97.0 $\pm0.8$ & \bf 97.0 $\pm0.8$ & \bf 97.5 $\pm0.7$ & \bf 97.5 $\pm0.9$\\
40922     &     96.6 $\pm0.4$ &     96.0 $\pm0.6$ & \bf 97.7 $\pm0.5$ & \bf 97.7 $\pm0.5$ & \bf 97.3 $\pm0.1$ & \bf 97.7 $\pm0.5$\\
1461      &     88.8 $\pm0.3$ &     88.8 $\pm0.2$ & \bf 89.4 $\pm0.1$ & \bf 89.4 $\pm0.1$ &     88.8 $\pm0.3$ & \bf 89.4 $\pm0.1$\\
41138     &     98.4 $\pm0.1$ &     98.4 $\pm0.1$ & \bf 98.9 $\pm0.1$ & \bf 98.9 $\pm0.1$ &     98.7 $\pm0.1$ & \bf 99.0 $\pm0.1$\\
42803     &     76.1 $\pm0.4$ &     75.6 $\pm0.5$ &     74.4 $\pm1.2$ & \bf 76.9 $\pm0.4$ &     76.2 $\pm0.3$ & \bf 76.9 $\pm0.4$\\
CIFAR-10  &     70.2 $\pm0.7$ &     70.4 $\pm0.4$ &     68.9 $\pm1.0$ & \bf 71.2 $\pm0.3$ &     66.9 $\pm1.2$ & \bf 71.6 $\pm0.4$\\
CIFAR-10S &     85.0 $\pm0.8$ &     84.3 $\pm0.5$ &     84.8 $\pm0.8$ & \bf 86.3 $\pm0.5$ &     85.9 $\pm0.5$ & \bf 86.4 $\pm0.6$\\
MNIST     &     82.6 $\pm0.8$ &     82.5 $\pm0.5$ &     82.0 $\pm0.9$ &     85.4 $\pm0.5$ &     80.5 $\pm1.2$ & \bf 87.1 $\pm0.4$\\

    \bottomrule
  \end{tabular}
  \caption{Benchmark results per dataset and sampling strategy. We show the average accuracy over 10 folds. Entropy (\emph{resp.} IWKMeans) has been omitted because their results were close to Confidence (\emph{resp.} WKMeans). Datasets are ordered to display patterns of dominance for samplers.}
  \label{benchmark-results}
\end{table}

\begin{figure}[pth]
    \centering
    \begin{subfigure}[b]{0.49\textwidth}
             \centering
             \includegraphics[width=\textwidth]{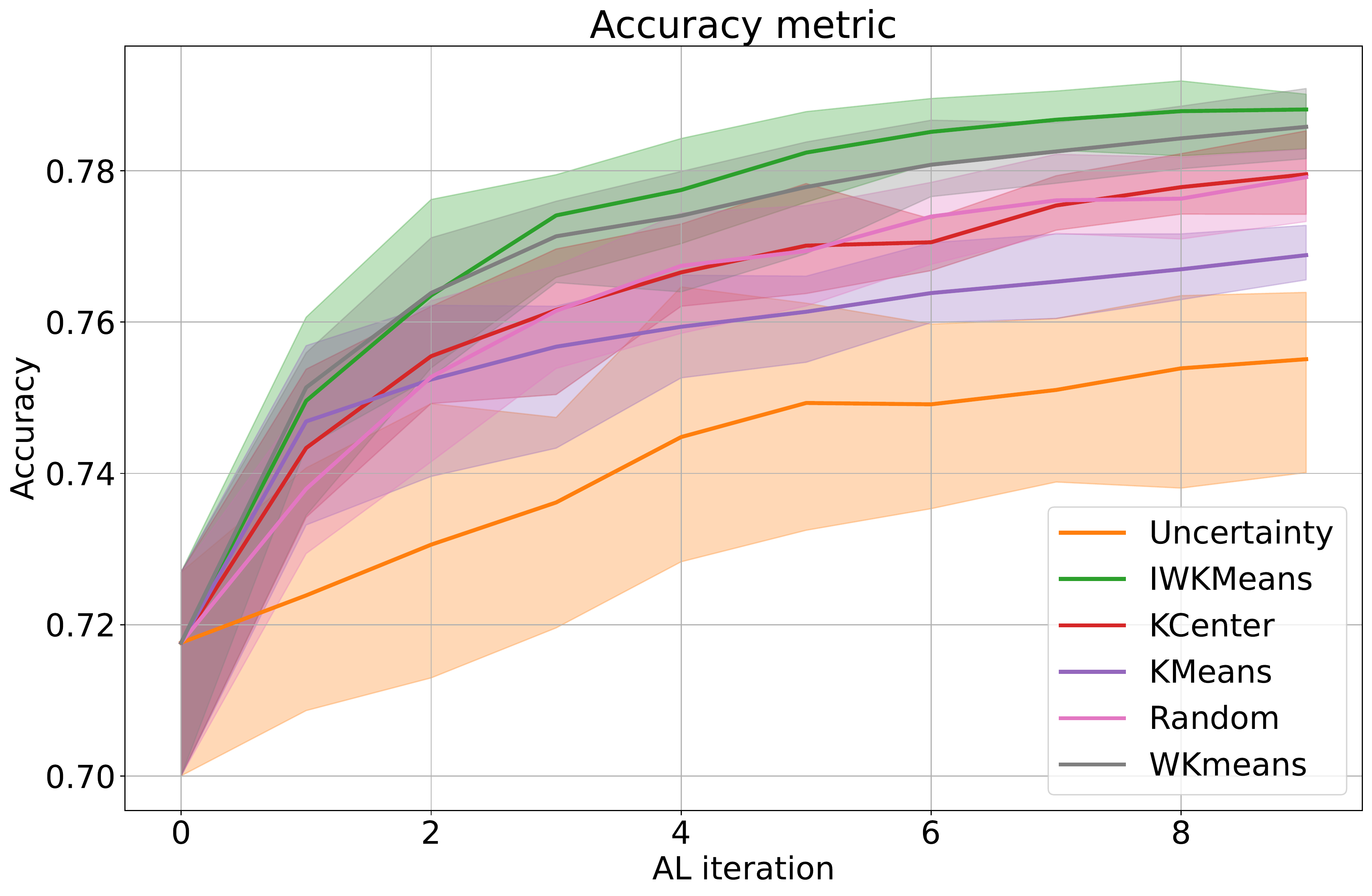}
         \end{subfigure}
         \hfill
         \begin{subfigure}[b]{0.49\textwidth}
             \centering
             \includegraphics[width=\textwidth]{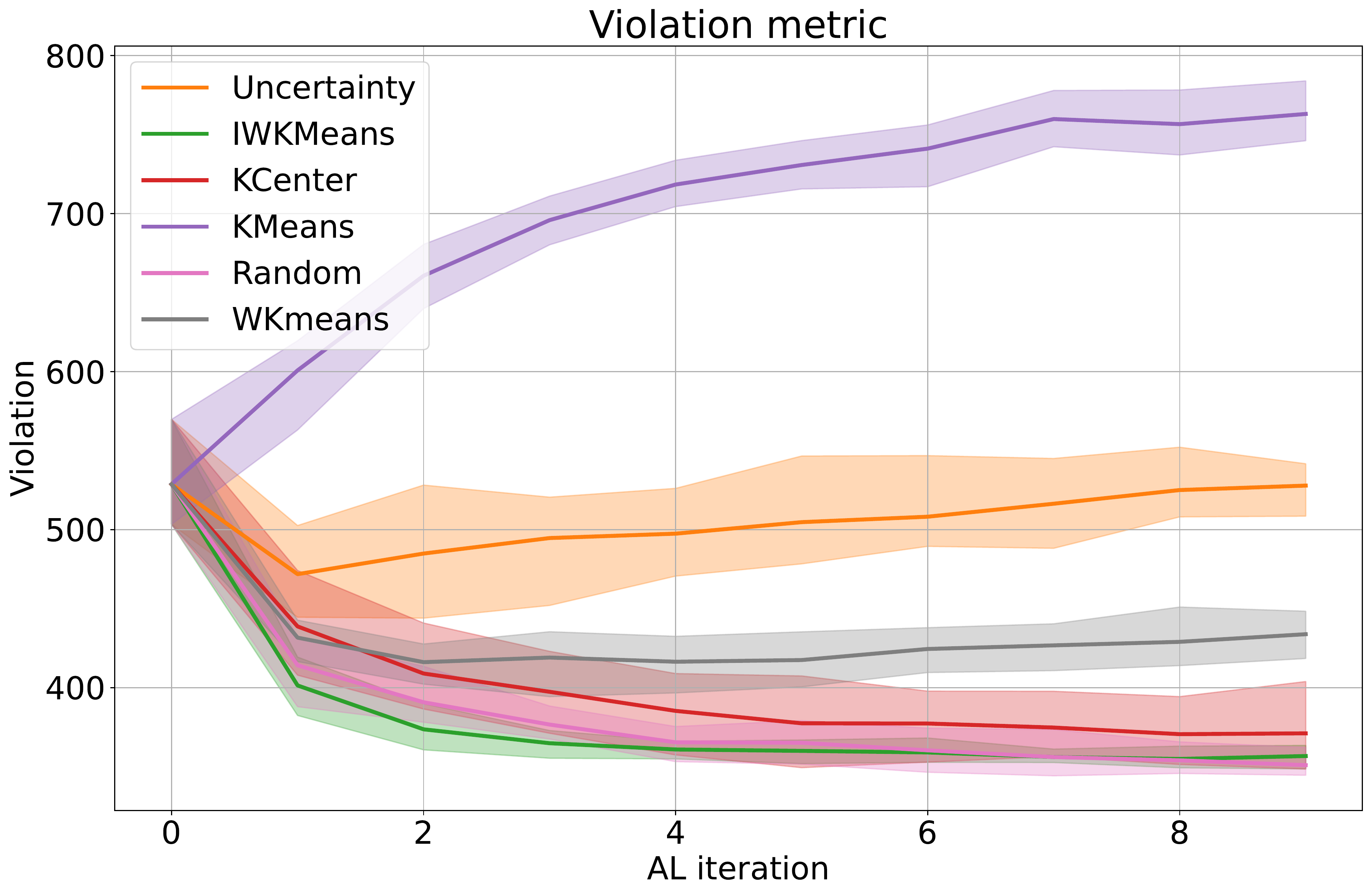}
         \end{subfigure}
         
    \captionsetup{justification=centering, margin=0cm}
    \caption{Results for dataset \#42803: Accuracy (left) and Violation (right). Note that the violation metrics is at 0 on the first iteration because the dataset is too small to compute them.}
    \label{fig:42803-accuracy-violation}
\end{figure}

\textbf{Best active learning strategy.} WKMeans and IWKMeans have similar performances and dominate the benchmark in terms of accuracy on all tasks, as observed in Table~\ref{benchmark-results} and in previous work~\cite{abraham2020rebuilding, abraham2021sample}. One notable difference is that IWKMeans has fewer violations than WKMeans which means that its training set is more representative of the test set, as observed in Figure \ref{fig:42803-accuracy-violation}. We expected this result as IWKMeans is designed to sample data more uniformly than WKMeans. If this does not impact accuracy, we could expect a different generalization power between the two models which advocates for adding a domain adaptation task in the future.

\begin{figure}[pth]
    \centering
    \begin{subfigure}[b]{0.49\textwidth}
             \centering
             \includegraphics[width=\textwidth]{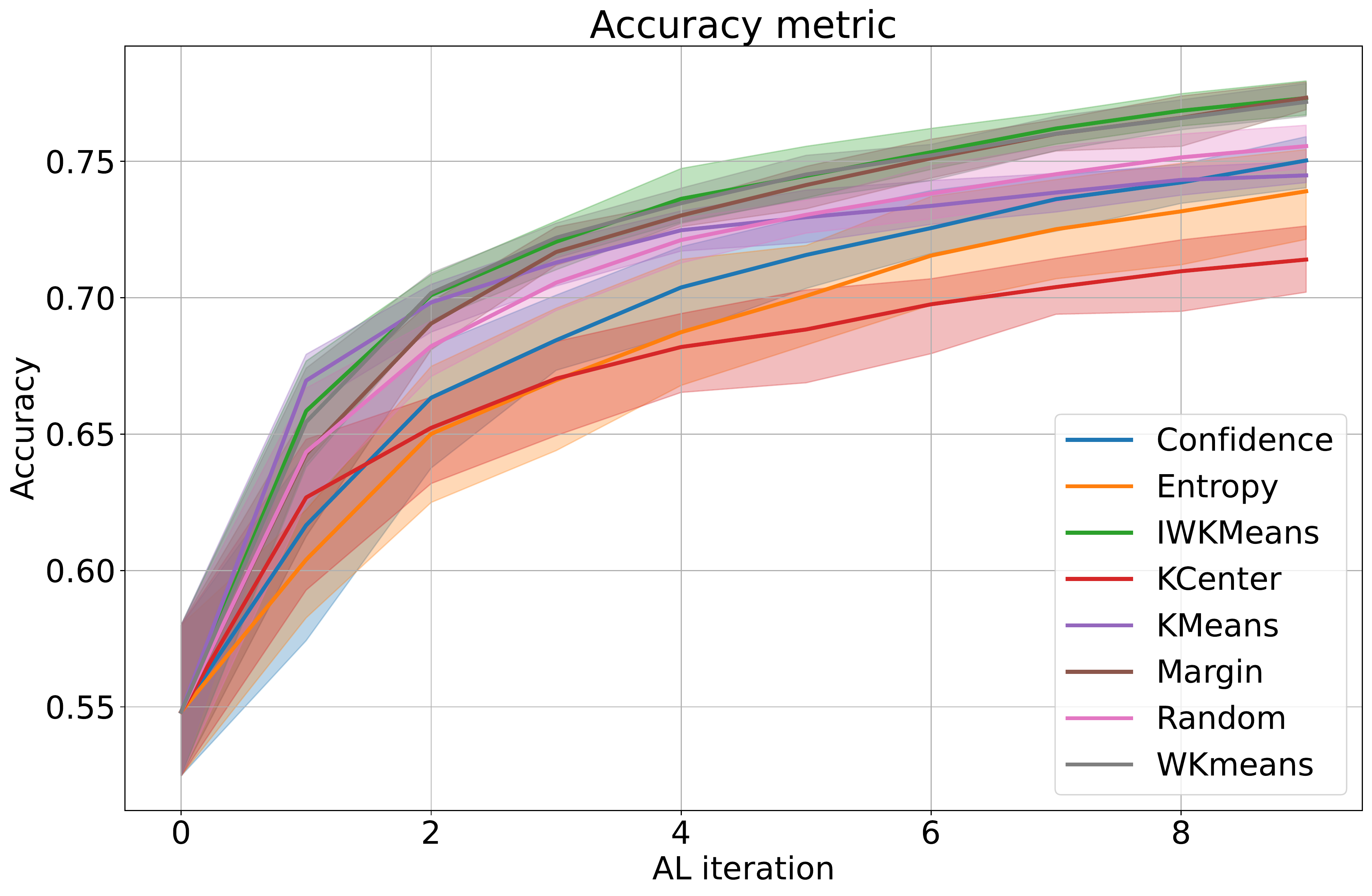}
         \end{subfigure}
         \hfill
         \begin{subfigure}[b]{0.49\textwidth}
             \centering
             \includegraphics[width=\textwidth]{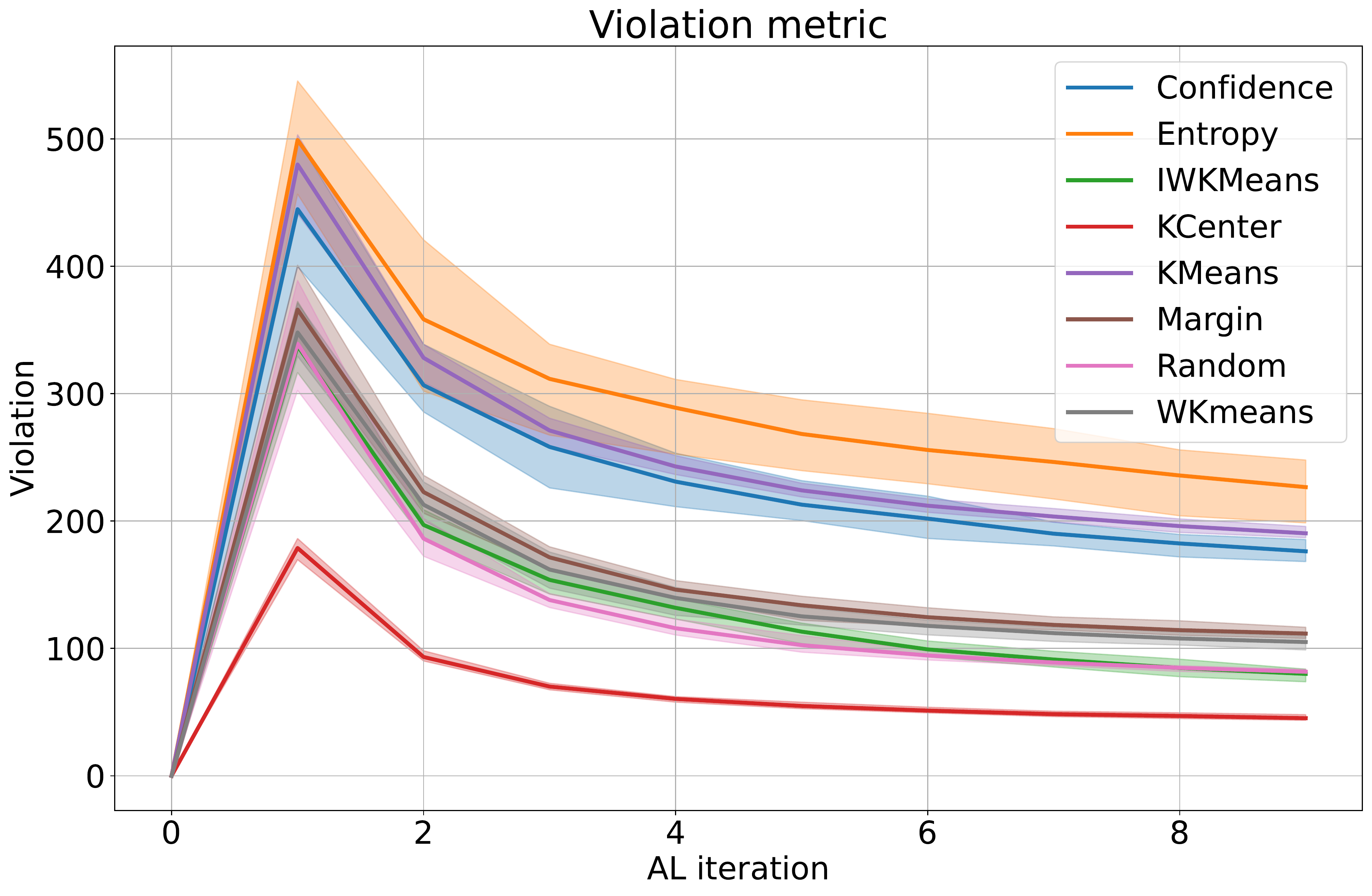}
         \end{subfigure}
         
    \captionsetup{justification=centering, margin=0cm}
    \caption{Results for dataset CIFAR-10: Accuracy (left) and Violation (right).}
    \label{fig:cifar10-accuracy-violation}
\end{figure}

\textbf{Uncertainty-based AL strategies.} As all uncertainty metrics have the same rank in binary classification, we resort to multi-class tasks to compare them. The only non-binary tabular classification task of our benchmark is \#42803. We observe that Confidence and Entropy strategies perform poorly, even worse than random. Looking at the metrics, we notice that most of the two strategies' values do not stand out except for higher violations for Confidence and Entropy as seen in Figure~\ref{fig:cifar10-accuracy-violation}. This means that the training set is very different from the test set, which may be due to those samplers focusing on noisy samples~\cite{abraham2021sample}. Margin reaffirms its dominance which explains why it is preferably used in many studies~\cite{zhdanov2019diverse}.

\begin{figure}[pth]
    \centering
    \begin{subfigure}[b]{0.49\textwidth}
             \centering
             \includegraphics[width=\textwidth]{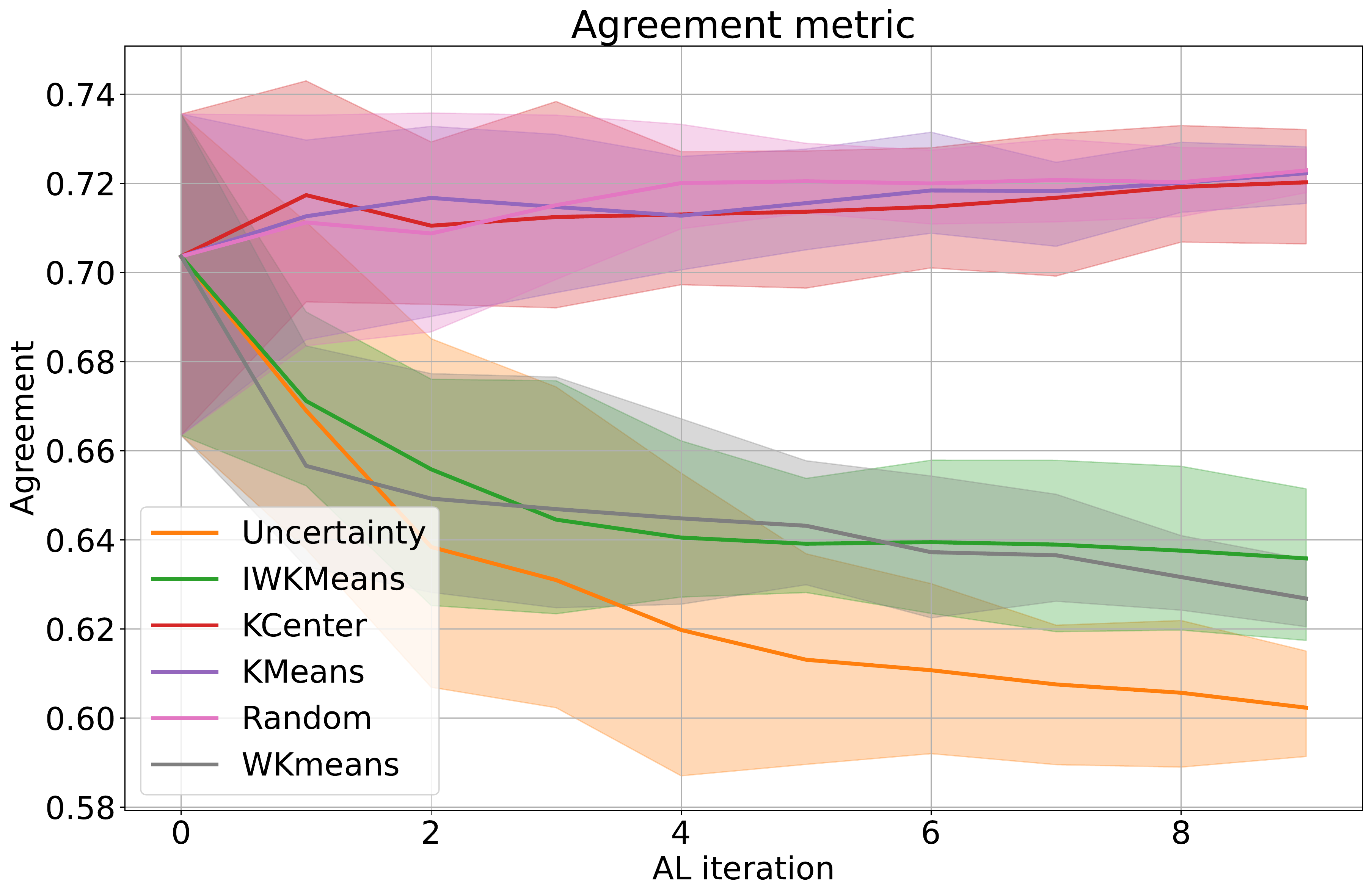}
         \end{subfigure}
         \hfill
         \begin{subfigure}[b]{0.49\textwidth}
             \centering
             \includegraphics[width=\textwidth]{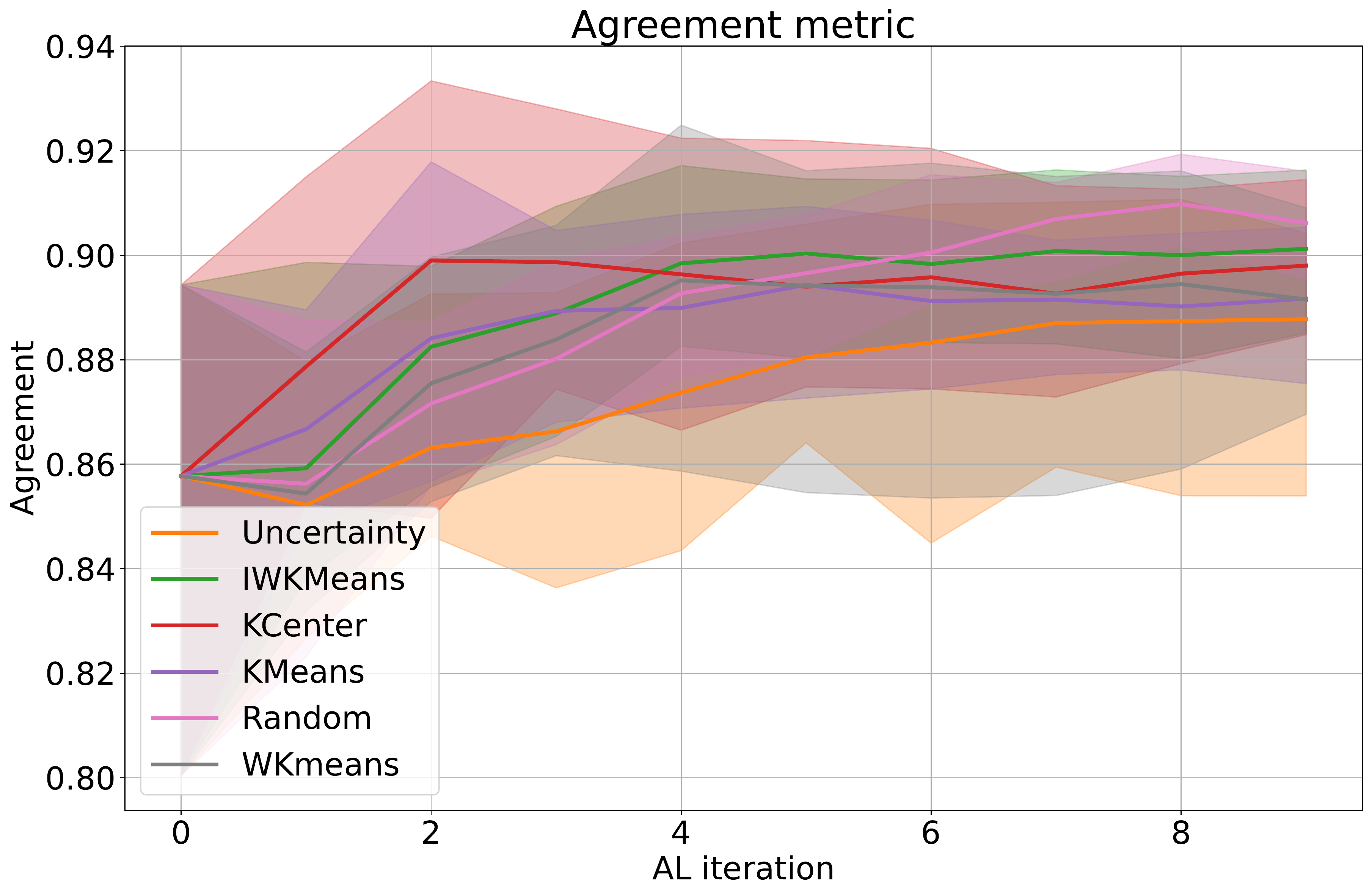}
         \end{subfigure}
         
    \captionsetup{justification=centering, margin=0cm}
    \caption{Agreements for datasets: \#42803 (left) and \#43439 (right).}
    \label{fig:agreement-42803-43439}
\end{figure}

\textbf{The importance of data representation.} IWKMeans and WKMeans are overall the best methods, but we observe a subset of methods on which uncertainty-based Margin is on par with them and another one where exploration-based KCenter reaches the same accuracy as displayed in Table~\ref{benchmark-results}. Surprisingly, the agreement metric seems to be a good indicator of which method is on par with the best. When all samplers have a similar agreement score, KCenter manages to reach the best accuracy. Conversely, when the agreement score of Margin is significantly below Random or Kcenter, Margin reaches the best performance. Figure~\ref{fig:agreement-42803-43439} illustrates the case of dominance of Margin on \#42803 and dominance of KCenter on \#43439. We hypothesize that the quality of the representation learned is responsible for this effect. When the distance in the representation space is inconsistent with the labels, diversity becomes ineffective or even counter-productive.

\begin{figure}[pth]
    \centering
    \begin{subfigure}[b]{0.49\textwidth}
             \centering
             \includegraphics[width=\textwidth]{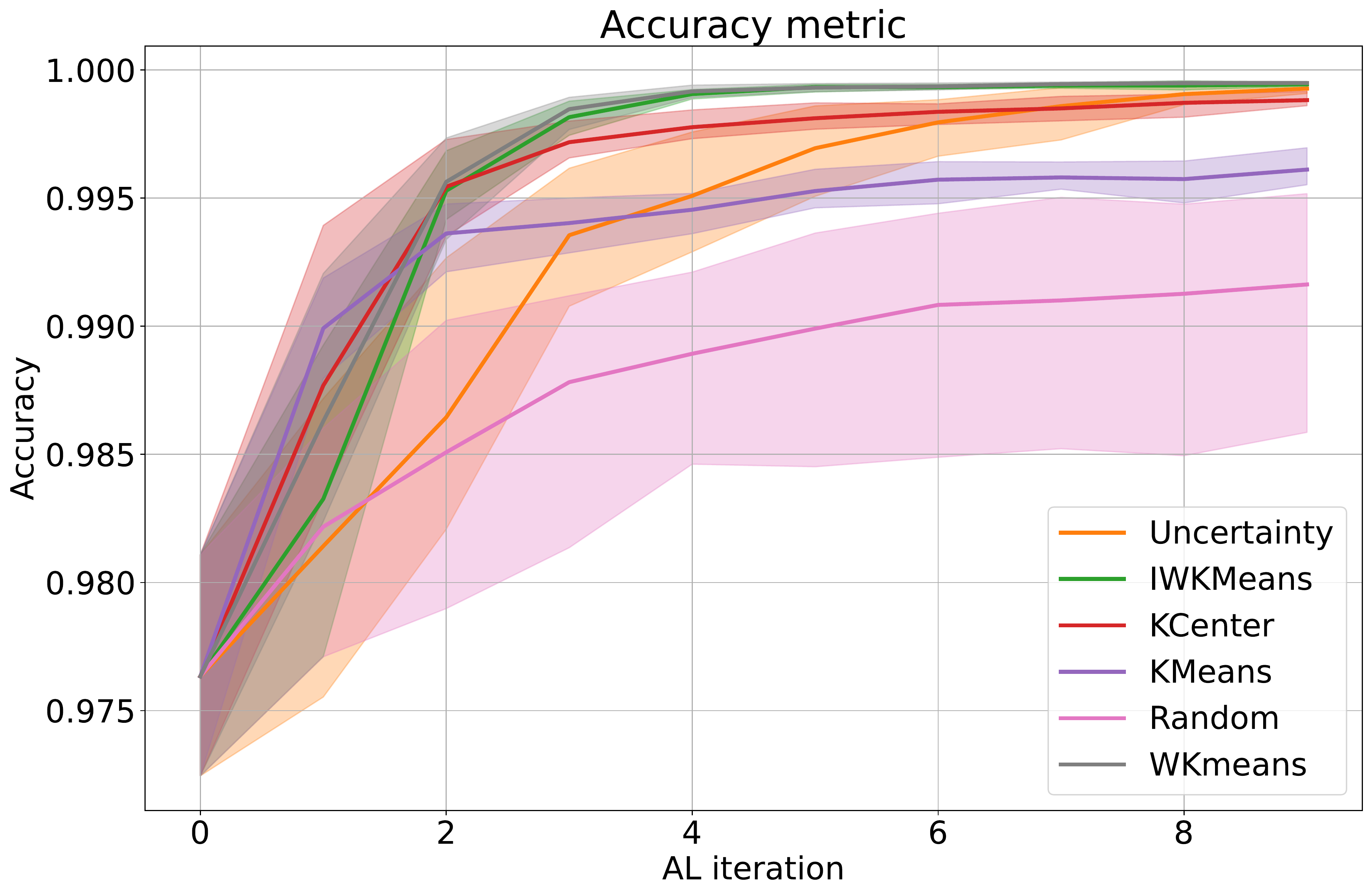}
         \end{subfigure}
         \hfill
         \begin{subfigure}[b]{0.49\textwidth}
             \centering
             \includegraphics[width=\textwidth]{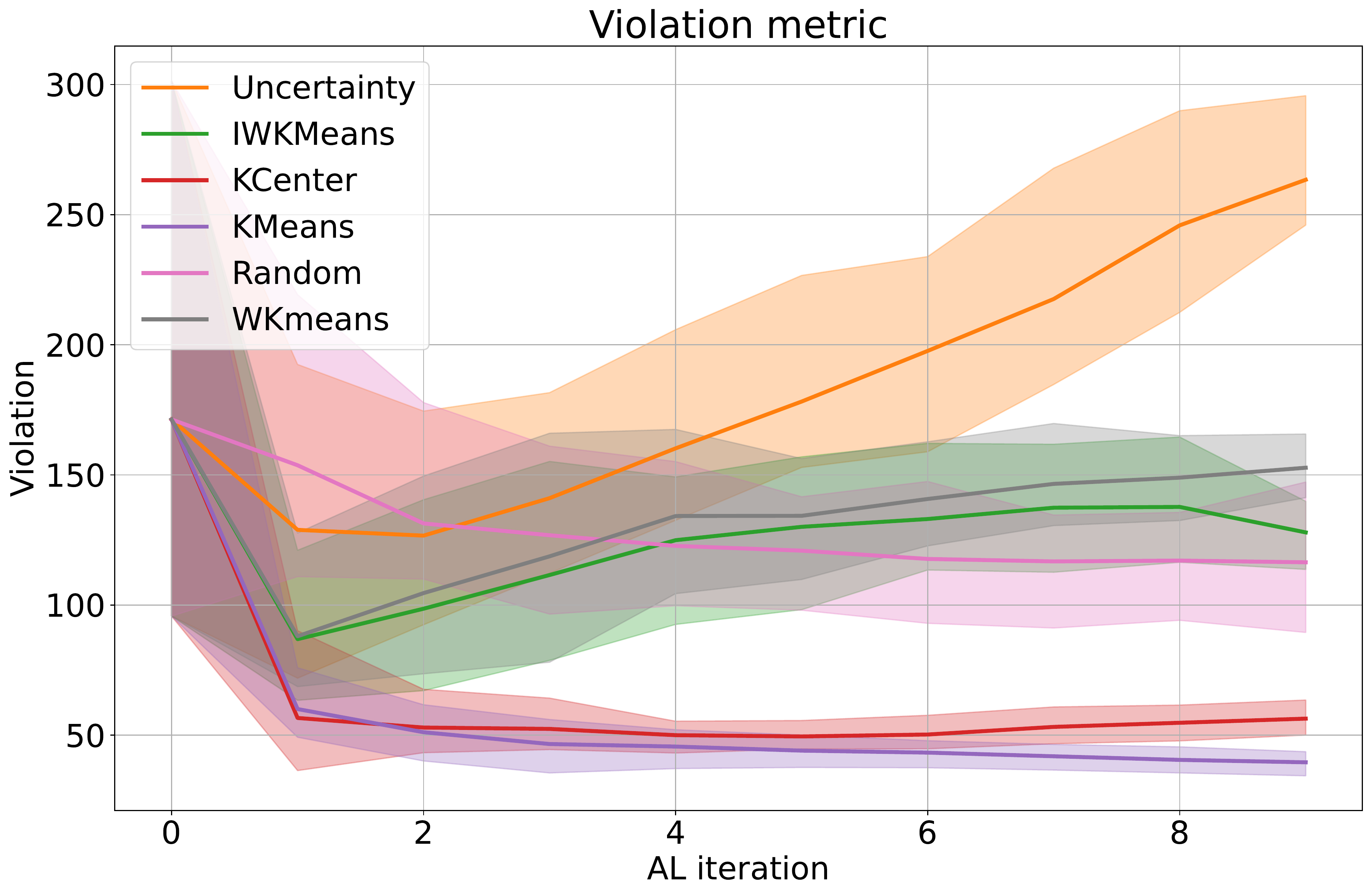}
         \end{subfigure}
         
    \captionsetup{justification=centering, margin=0cm}
    \caption{Results for dataset \#1502: Accuracy (left) and Violation (right).}
    \label{fig:1502-accuracy-violation}
\end{figure}

\textbf{The peculiar case of \#1502.} Task \#1502 is peculiar since we considered it tabular, but its three features are an image's red, green, and blue channels. It is the only task where KMeans has better accuracy than random. More than that, we observe two regimes in this experiment. KMeans and KCenter-Greedy, two purely exploratory techniques, dominate the two first iterations of the experiment. After that, they plateau at a suboptimal accuracy, while uncertainty based-methods take the lead. We hypothesize that at iteration 1, the training set is too small for the model to produce meaningful uncertainty scores. WKMeans, which combines uncertainty and exploration, manages to take the best of both worlds. This unusual behavior could be correlated to the unique pattern shown in the violations metric where KMeans minimize this score while WKMeans keeps it increasing, as displayed in Figure~\ref{fig:1502-accuracy-violation}. Unfortunately, we cannot draw a conclusion from one task, and we hope that adding further tasks could help us reproduce this behavior and understand it better.

         

\section{Limitations}

We have limited ourselves to OpenML datasets and the most common image ones for this proof of concept. In the future, we plan to explore other data sources, such as Kaggle, and other modalities or tasks, such as text and object detection. We could also explore other models and settings, such as different batch sizes, to observe their effect on overall performance. Given the high computational cost of the benchmark, we have set aside very costly methods such as BADGE~\cite{ash2019deep} or BatchBALD, but we plan to add them in the near future.

\section{Conclusion}

This first version of OpenAL proves the value of comparing methods on fixed predefined tasks. Although the global outcome that the more sophisticated methods dominate the others was expected, the systematic monitoring and analysis of the metrics helped dig into the results. We believe that our benchmark is a first step towards helping the practitioners to be more confident in their choice of samplers.

By ensuring a complete reproducibility of the results, we also allow strategy developer to test their method against our references quickly. Thanks to the metrics, they can also understand faster why their method may fail on a peculiar dataset and why other methods perform better. For example, we have highlighted that diversity in active learning strategies is as good as the data representation on which it relies.


In the end, we hope this benchmark will accelerate research in active learning and facilitate its adoption in industrial contexts.

\bibliography{main}

\begin{thebibliography}{10}

\bibitem{abraham2020rebuilding}
Alexandre Abraham and L{\'e}o Dreyfus-Schmidt.
\newblock Rebuilding trust in active learning with actionable metrics.
\newblock {\em 2020 IEEE International Conference on Data Mining Workshops
  (ICDMW)}, 2020.

\bibitem{abraham2021sample}
Alexandre Abraham and L{\'e}o Dreyfus-Schmidt.
\newblock Sample noise impact on active learning.
\newblock {\em IAL 2021 workshop, ECML PKDD}, 2021.

\bibitem{ash2019deep}
Jordan~T Ash, Chicheng Zhang, Akshay Krishnamurthy, John Langford, and Alekh
  Agarwal.
\newblock Deep batch active learning by diverse, uncertain gradient lower
  bounds.
\newblock {\em arXiv preprint arXiv:1906.03671}, 2019.

\bibitem{chen2020simple}
Ting Chen, Simon Kornblith, Mohammad Norouzi, and Geoffrey Hinton.
\newblock A simple framework for contrastive learning of visual
  representations.
\newblock In {\em International conference on machine learning}, pages
  1597--1607. PMLR, 2020.

\bibitem{dietterich1998approximate}
Thomas~G Dietterich.
\newblock Approximate statistical tests for comparing supervised classification
  learning algorithms.
\newblock {\em Neural computation}, 10(7):1895--1923, 1998.

\bibitem{fariha2021conformance}
Anna Fariha, Ashish Tiwari, Arjun Radhakrishna, Sumit Gulwani, and Alexandra
  Meliou.
\newblock Conformance constraint discovery: Measuring trust in data-driven
  systems.
\newblock In {\em Proceedings of the 2021 International Conference on
  Management of Data}, pages 499--512, 2021.

\bibitem{feurer2021openml}
Matthias Feurer, Jan~N Van~Rijn, Arlind Kadra, Pieter Gijsbers, Neeratyoy
  Mallik, Sahithya Ravi, Andreas M{\"u}ller, Joaquin Vanschoren, and Frank
  Hutter.
\newblock Openml-python: an extensible python api for openml.
\newblock {\em The Journal of Machine Learning Research}, 22(1):4573--4577,
  2021.

\bibitem{kirsch2019batchbald}
Andreas Kirsch, Joost Van~Amersfoort, and Yarin Gal.
\newblock Batchbald: Efficient and diverse batch acquisition for deep bayesian
  active learning.
\newblock {\em arXiv preprint arXiv:1906.08158}, 2019.

\bibitem{kottke2017challenges}
Daniel Kottke, Adrian Calma, Denis Huseljic, GM~Krempl, and Bernhard Sick.
\newblock Challenges of reliable, realistic and comparable active learning
  evaluation.
\newblock In {\em Proceedings of the Workshop and Tutorial on Interactive
  Adaptive Learning}, pages 2--14, 2017.

\bibitem{kottke2019limitations}
Daniel Kottke, Jim Schellinger, Denis Huseljic, and Bernhard Sick.
\newblock Limitations of assessing active learning performance at runtime.
\newblock {\em arXiv preprint arXiv:1901.10338}, 2019.

\bibitem{limberg2020beyond}
Christian Limberg, Heiko Wersing, and Helge Ritter.
\newblock Beyond cross-validation—accuracy estimation for incremental and
  active learning models.
\newblock {\em Machine Learning and Knowledge Extraction}, 2(3):327--346, 2020.

\bibitem{lowell2018practical}
David Lowell, Zachary~C Lipton, and Byron~C Wallace.
\newblock Practical obstacles to deploying active learning.
\newblock {\em arXiv preprint arXiv:1807.04801}, 2018.

\bibitem{moustapha2022active}
Maliki Moustapha, Stefano Marelli, and Bruno Sudret.
\newblock Active learning for structural reliability: Survey, general framework
  and benchmark.
\newblock {\em Structural Safety}, 96:102174, 2022.

\bibitem{munjal2020towards}
Prateek Munjal, Nasir Hayat, Munawar Hayat, Jamshid Sourati, and Shadab Khan.
\newblock Towards robust and reproducible active learning using neural
  networks.
\newblock {\em arXiv}, pages arXiv--2002, 2020.

\bibitem{ren2021survey}
Pengzhen Ren, Yun Xiao, Xiaojun Chang, Po-Yao Huang, Zhihui Li, Brij~B Gupta,
  Xiaojiang Chen, and Xin Wang.
\newblock A survey of deep active learning.
\newblock {\em ACM computing surveys (CSUR)}, 54(9):1--40, 2021.

\bibitem{sener2017active}
Ozan Sener and Silvio Savarese.
\newblock Active learning for convolutional neural networks: A core-set
  approach.
\newblock {\em arXiv preprint arXiv:1708.00489}, 2017.

\bibitem{settles2009active}
Burr Settles.
\newblock Active learning literature survey.
\newblock Technical report, Department of Computer Sciences, University of
  Wisconsin-Madison, 2009.

\bibitem{sinha2019variational}
Samarth Sinha, Sayna Ebrahimi, and Trevor Darrell.
\newblock Variational adversarial active learning.
\newblock In {\em Proceedings of the IEEE/CVF International Conference on
  Computer Vision}, pages 5972--5981, 2019.

\bibitem{trittenbach2021overview}
Holger Trittenbach, Adrian Englhardt, and Klemens B{\"o}hm.
\newblock An overview and a benchmark of active learning for outlier detection
  with one-class classifiers.
\newblock {\em Expert Systems with Applications}, 168:114372, 2021.

\bibitem{vanschoren2014openml}
Joaquin Vanschoren, Jan~N Van~Rijn, Bernd Bischl, and Luis Torgo.
\newblock Openml: networked science in machine learning.
\newblock {\em ACM SIGKDD Explorations Newsletter}, 15(2):49--60, 2014.

\bibitem{wiens2010active}
Jenna Wiens and John Guttag.
\newblock Active learning applied to patient-adaptive heartbeat classification.
\newblock {\em Advances in neural information processing systems}, 23, 2010.

\bibitem{yang2018benchmark}
Yazhou Yang and Marco Loog.
\newblock A benchmark and comparison of active learning for logistic
  regression.
\newblock {\em Pattern Recognition}, 83:401--415, 2018.

\bibitem{zhdanov2019diverse}
Fedor Zhdanov.
\newblock Diverse mini-batch active learning.
\newblock {\em arXiv preprint arXiv:1901.05954}, 2019.

\end{thebibliography}

\newpage
\section*{Appendix}
\appendix

\counterwithin{figure}{section}
\section{Tabular dataset metrics}

\newcommand\subfigwidth{0.4}
\foreach \datasetid in {1461, 1471, 1502, 1590, 42395, 43439, 43551, 42803, 41162, 40922, 41138}
    {
    \begin{figure}
        \centering
             \begin{subfigure}[b]{\subfigwidth\textwidth}
                 \centering
                 \includegraphics[width=\textwidth]{figures/\datasetid/plot-Accuracy.pdf}
                 \caption{Accuracy}
             \end{subfigure}
             \begin{subfigure}[b]{\subfigwidth\textwidth}
                 \centering
                 \includegraphics[width=\textwidth]{figures/\datasetid/plot-Agreement.pdf}
                 \caption{Agreement}
             \end{subfigure}
             \begin{subfigure}[b]{\subfigwidth\textwidth}
                 \centering
                 \includegraphics[width=\textwidth]{figures/\datasetid/plot-Contradictions.pdf}
                 \caption{Contradictions}
             \end{subfigure}
             \begin{subfigure}[b]{\subfigwidth\textwidth}
                 \centering
                 \includegraphics[width=\textwidth]{figures/\datasetid/plot-F-Score.pdf}
                 \caption{F-Score}
             \end{subfigure}
             \begin{subfigure}[b]{\subfigwidth\textwidth}
                 \centering
                 \includegraphics[width=\textwidth]{figures/\datasetid/plot-Hard-Exploration.pdf}
                 \caption{Hard-Exploration}
             \end{subfigure}
             \begin{subfigure}[b]{\subfigwidth\textwidth}
                 \centering
                 \includegraphics[width=\textwidth]{figures/\datasetid/plot-Top-Exploration.pdf}
                 \caption{Top-Exploration}
             \end{subfigure}
             \begin{subfigure}[b]{\subfigwidth\textwidth}
                 \centering
                 \includegraphics[width=\textwidth]{figures/\datasetid/plot-Violation.pdf}
                 \caption{Violation}
             \end{subfigure}
        \captionsetup{justification=centering, margin=0cm}
        \caption{Samplers performances on \datasetid.}
    \end{figure}
    }

\newpage
\counterwithin{figure}{section}
\section{Image dataset metrics}

\foreach \datasetid in {mnist, cifar10, cifar10-simclr}
    {
    \begin{figure}
        \centering
             \begin{subfigure}[b]{\subfigwidth\textwidth}
                 \centering
                 \includegraphics[width=\textwidth]{figures/\datasetid/plot-Accuracy.pdf}
                 \caption{Accuracy}
             \end{subfigure}
             \begin{subfigure}[b]{\subfigwidth\textwidth}
                 \centering
                 \includegraphics[width=\textwidth]{figures/\datasetid/plot-Agreement.pdf}
                 \caption{Agreement}
             \end{subfigure}
             \begin{subfigure}[b]{\subfigwidth\textwidth}
                 \centering
                 \includegraphics[width=\textwidth]{figures/\datasetid/plot-Contradictions.pdf}
                 \caption{Contradictions}
             \end{subfigure}
             \begin{subfigure}[b]{\subfigwidth\textwidth}
                 \centering
                 \includegraphics[width=\textwidth]{figures/\datasetid/plot-Hard-Exploration.pdf}
                 \caption{Hard-Exploration}
             \end{subfigure}
             \begin{subfigure}[b]{\subfigwidth\textwidth}
                 \centering
                 \includegraphics[width=\textwidth]{figures/\datasetid/plot-Top-Exploration.pdf}
                 \caption{Top-Exploration}
             \end{subfigure}
             \begin{subfigure}[b]{\subfigwidth\textwidth}
                 \centering
                 \includegraphics[width=\textwidth]{figures/\datasetid/plot-Violation.pdf}
                 \caption{Violation}
             \end{subfigure}
        \captionsetup{justification=centering, margin=0cm}
        \caption{Samplers performances on \datasetid.}
    \end{figure}
    }

\end{document}